\pdfoutput=1

\documentclass[11pt]{article}
\usepackage{authblk}
\usepackage[]{EMNLP2022}

\usepackage{times}
\usepackage{latexsym}
\usepackage{makecell}
\usepackage{multirow}

\usepackage{amssymb}
\usepackage{pifont}
\newcommand{\cmark}{\ding{51}}%
\newcommand{\xmark}{\ding{55}}%

\usepackage[T1]{fontenc}

\usepackage[utf8]{inputenc}

\usepackage{microtype}

\usepackage{inconsolata}

\usepackage{tikz}
\usepackage{amsmath}

\newcommand{\shortrightarrow}[1]{%
\parbox{#1}{\tikz{\draw[->](0,0)--(#1,0);}}
}

\makeatletter
\newcommand\blfootnote[1]{%
  \begingroup
  \renewcommand{\@makefntext}[1]{\noindent\makebox[1.8em][r]#1}
  \renewcommand\thefootnote{}\footnote{#1}%
  \addtocounter{footnote}{-1}%
  \endgroup
}
\makeatother

\title{DATScore: Evaluating Translation with Data Augmented Translations}

\author[1]{Moussa Kamal Eddine}
\author[2]{Guokan Shang}
\author[1,3]{Michalis Vazirgiannis}
\affil[1]{\'Ecole Polytechnique, $^\mathrm{2}$Linagora, $^\mathrm{3}$AUEB}

\begin{document}
\maketitle

\begin{abstract}
The rapid development of large pretrained language models has revolutionized not only the field of Natural Language Generation (NLG) but also its evaluation. 
Inspired by the recent work of BARTScore: a metric leveraging the BART language model to evaluate the quality of generated text from various aspects, we introduce DATScore.
DATScore uses data augmentation techniques to improve the evaluation of machine translation. 
Our main finding is that introducing data augmented translations of the source and reference texts is greatly helpful in evaluating the quality of the generated translation.
We also propose two novel score averaging and term weighting strategies to improve the original score computing process of BARTScore.
Experimental results on WMT show that DATScore correlates better with human meta-evaluations than the other recent state-of-the-art metrics, especially for low-resource languages. 
Ablation studies demonstrate the value added by our new scoring strategies.
Moreover, we report in our extended experiments the performance of DATScore on 3 NLG tasks other than translation.
Code is publicly available\footnote{https://github.com/moussaKam/dat\_score}.
\end{abstract}

\section{Introduction}
Massive pretrained language models have brought significant improvement to NLG tasks \citep{lewis-etal-2020-bart}.
Recent systems can even generate texts of higher quality than human-annotated ones \cite{peyrard-2019-studying}.
At the same time,  standard metrics, such as BLEU \citep{papineni-etal-2002-bleu} and ROUGE \citep{lin-2004-rouge}, for translation and summarization respectively, have not evolved for the past two decades \citep{bhandari-etal-2020-evaluating}.
These metrics rely on surface lexicographic matches, making them particularly unsuitable for evaluating modern systems operating with embeddings at the semantic level that often generate paraphrases \citep{ng-abrecht-2015-better}.
To address this issue, many metrics have been proposed \citep{sai2022survey}, but none of them were widely adopted until the release of BERTSore \citep{zhang2019bertscore} and MoverScore \citep{zhao-etal-2019-moverscore}.
These metrics take advantage of large pretrained language models like BERT \citep{devlin-etal-2019-bert}, which are now being used in nearly all NLP tasks \citep{qiu2020pre,min2021recent}.

In this work, we focus on the task of evaluating machine translation.
We propose an extension of BARTScore \citep{yuan2021bartscore}, a recent metric exploiting the BART seq2seq language model \citep{lewis-etal-2020-bart} to evaluate the quality of generated text from various aspects.
BARTScore covers four evaluation facets: Faithfulness, Precision, Recall, and F-score, derived from different generation directions between the \textit{source} text, the \textit{hypothesis} (the text generated by a system given the source), and the \textit{reference} (the reference text for the generation, often provided by human annotators).
The scores are obtained by pairing the three entities differently at the input or the output side of a trained seq2seq model for fetching conditional generation probabilities.

Based on BARTScore, and motivated by the general idea and positive effect of data augmentation techniques, we found that adding augmented, translated copies of the source and reference texts in BARTScore, can greatly help evaluate the quality of the hypothesis translation.
We also propose two novel score averaging and term weighting strategies to improve the original score computing process of BARTScore.
Results and ablation studies show that our metric DATScore (Data Augmented Translation Score) outperforms the other recent state-of-the-art metrics, and our new scoring strategies are effective.
Moreover, the performance of DATScore is also reported on three other NLG tasks than translation: data-to-text, summarization, and image captioning.

To the best of our knowledge, no prior work has been done on leveraging data augmentation techniques for untrained NLG evaluation metrics.
Our work will help fill this gap.
Our contributions include: \\
1) Inspired by BARTScore, we developed DATScore, incorporating augmented data translated from the source and reference texts.
DATScore is an untrained and unsupervised translation evaluation metric that offers a larger performance boost in evaluating low-resource language generation.
In contrast to other widely adopted metrics, DATScore can efficiently incorporate both the source and reference texts in the evaluation.
\\
2) We introduced a novel one-vs-rest method to average the scores for different generation directions with different weights, which improves over the simple arithmetic averaging method used in BARTScore. \\
3) We proposed a novel entropy-based scheme for weighting the target generated terms so that higher informative tokens receive more importance in accounting for the score, which outperforms the naive uniform weighting employed in BARTScore.

\section{Related work}

\subsection{Translation evaluation metrics} \label{subsec:metrics}

BLEU \citep{papineni-etal-2002-bleu} is the de facto metric for evaluating machine translation. It simply calculates $n$-gram matching between the reference and the hypothesis using precision scores with a brevity penalty.
METEOR \citep{banerjee-lavie-2005-meteor} was developed to address two drawbacks of BLEU. It is F-score based (thus taking recall into account) and allows for a more relaxed matching, based on three forms: extract unigram, stemmed word, and synonym with WordNet \citep{miller-1994-wordnet}.
Apart from the above word-based metrics, some approaches operate at the character level. For example, chrF \citep{popovic-2015-chrf} computes the overall precision and recall over the character $n$-grams with various values of $n$.
More recently, static word embeddings \citep{mikolov2013efficient} have enabled capturing the semantic similarity between two texts possible, of what the historical metrics are incapable.
Several metrics have been proposed to incorporate word vectors.
For example, MEANT 2.0 \citep{lo-2017-meant} evaluates translation adequacy by measuring the similarity of the semantic frames and their role fillers between the human and machine translations.

Lately, pretrained language models have become popular, because they provide context-dependent embeddings.
This proved beneficial to all NLP tasks, but also to evaluation metrics.
For example, using a modified version of the Word Mover's Distance \citep{kusner2015word}, the Sentence Mover's Similarity \citep{clark-etal-2019-sentence} measures the minimum cost of transforming one text into the other as the evaluation score, where sentences are represented as the average of their ELMo word embeddings \citep{peters-etal-2018-deep}.
\textsc{BERTr} \citep{mathur-etal-2019-putting} computes approximate recall based on the pairwise cosine similarity between the BERT word embeddings \citep{devlin-etal-2019-bert} of two translations.
UniTE \citep{wan-etal-2022-unite} proposes a unified framework for modeling three evaluation prototypes: estimating the quality of the translation hypothesis by comparing it with reference-only, source-only, or source-reference-combined data.
UniTE is built upon XLM-R multilingual language model \citep{conneau-etal-2020-unsupervised}. 

Among several alternatives, BERTSore \citep{zhang2019bertscore} and MoverScore \citep{zhao-etal-2019-moverscore} have received more attention, and have been adopted for reporting results in recent NLG publications \citep{lin-etal-2022-roles,weston-etal-2022-generative}. They both are unsupervised, general-purpose metrics and leverage BERT-like language models, however, with one difference lying in the similarity function for matching the two sequence representations.
BERTScore greedily matches each token from one sequence to the single most similar token in the other sequence, in terms of the cosine similarity of their token embeddings. While MoverScore conducts soft one-to-many matching using an $n$-gram generalization of the Word Mover's Distance \citep{kusner2015word}.

Finally, the work closely related to ours is BARTScore \citep{yuan2021bartscore}. Unlike all the above metrics trying to match tokens or their embeddings, BARTScore proposes a novel conceptual view. It treats the evaluation of generated text as a text generation problem, with the help of a pretrained seq2seq model BART \citep{lewis-etal-2020-bart}.
At the time of writing, this metric represents the state-of-the-art in the NLG evaluation.
We will
provide more details about it in Section \ref{sec:datscore}.

\subsection{Data augmentation}
As deep learning models are often heavily reliant on large amounts of training data, a common attempt to get around the data scarcity problem is by applying data augmentation techniques \citep{shorten2019survey}.
These techniques increase the size of the training set by making slightly modified copies of already-existing instances or by creating new, synthetic ones.
Such augmented data have proven to be beneficial to the training of models in a wide variety of contexts, from computer vision \citep{shorten2019survey} to speech recognition \citep{bird2020overcoming}, to NLP \citep{feng-etal-2021-survey}, as it acts as a regularizer and helps reduce overfitting \citep{krizhevsky2012imagenet}. 
For dealing with textual data, a suite of augmentation techniques exists.
To name only a few, backtranslation \citep{sennrich-etal-2016-improving} translates a text into an intermediate language and then back into the original language, as a way of paraphrasing the initial text.
Contextual augmentation \citep{kobayashi-2018-contextual} generates augmented samples by randomly replacing words with others drawn following the in-context word distribution of a recurrent language model.
SeqMix method \citep{guo-etal-2020-sequence} creates synthetic examples by softly mixing parts of two sentences via a convex combination.

Data augmentation has also been applied to the field of NLG evaluation metrics. BLEURT \citep{sellam-etal-2020-bleurt} is a supervised metric, i.e., it requires to be finetuned on human meta-evaluations. Before finetuning, BLEURT creates an augmented synthetic dataset by perturbing Wikipedia sentences with BERT mask-filling, backtranslation, and random word dropping techniques.
The data are then annotated with some automatic numerical and categorical signals as pretraining labels.
FrugalScore \citep{kamal-eddine-etal-2022-frugalscore} proposes the first knowledge distillation approach for NLG evaluation metrics, to alleviate the significant requirement of computational resources by the heavy metrics based on large pretrained language models (e.g., BERTScore and MoverScore).
Unlike BLEURT, it is purely trained on a synthetic dataset consisting of pairs of more or less related sentences, created via various data augmentation techniques (e.g., paraphrasing with backtranslation, perturbation then denoising, etc.).
The sentence pairs for training the student model are annotated with scores given by the metrics to be learned.\\

\noindent\textbf{Differences.}
Note that BLEURT and FrugalScore use augmented data to train their parameterized metric models, while our DATScore is an untrained and unsupervised metric not requiring human judgments for training and using augmented translation for the sole purpose of scoring.

\section{DATScore} \label{sec:datscore}
As mentioned in Subsection \ref{subsec:metrics}, BARTScore is not based on matching tokens nor their embeddings as the other evaluation metrics.
Instead, it uses a novel approach by framing the evaluation of generated text as a text generation problem.
Assuming first a pretrained seq2seq model is ``ideal'' (e.g., BART), BARTScore directly uses the model's conditional probability of generating a provided target text $Y$ given a provided input text $X$, as the evaluation score of the generation direction $X\rightarrow Y$.
For example, $Y$ corresponds to a translation hypothesis generated by any system, and $X$ is the reference.
If $Y$ is of high quality, then by providing the pair to the pretrained BART model, the estimated conditional generation probability (evaluation score) $P(Y|X)$ should be high.

\begin{figure}[t]
    \centering
    \includegraphics[scale=0.6]{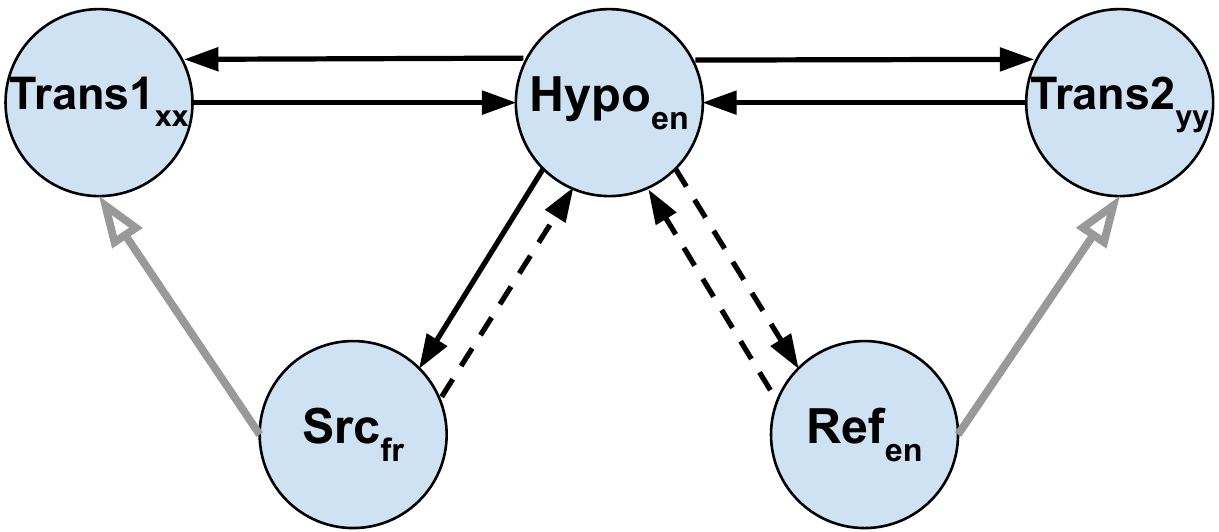}
	\caption{Dashed arrows denote the generation directions covered by BARTScore. Solid black arrows indicate our newly introduced directions for calculating DATScore of the example \textit{hypothesis} in English (Hypo$_{en}$). Trans1$_{xx}$ and Trans2$_{yy}$ represent data \textit{augmented translations} in any languages xx and yy, obtained by applying a translation model (grey arrows) to the example \textit{source} in French (Src$_{fr}$) and example \textit{reference} in English (Ref$_{en}$), respectively.}
    \label{fig:model}
\end{figure}

Therefore, by placing differently the \textit{source} (Src), the \textit{reference} (Ref), and the \textit{hypothesis} (Hypo) in pair at the input or the output side of the trained seq2seq model for fetching conditional generation probabilities, BARTScore considers three different generation directions illustrated as dashed arrows in Figure \ref{fig:model}.
The conditional probabilities associated with the directions are denoted as: Precision (\textit{Ref$\rightarrow$Hypo}), Recall (\textit{Hypo$\rightarrow$Ref}) and Faithfulness\footnote{BART being a monolingual model, faithfulness is only relevant in the context of abstractive summarization, and its corresponding direction cannot be applied to machine translation evaluation.} (\textit{Src$\rightarrow$Hypo}). Additionally, an F-score, the arithmetic average of Precision and Recall.


The score (conditional probability) for the generation direction from a source sequence $X=\{x_t\}_{t=1}^n$ to a target sequence $Y=\{y_t\}_{t=1}^m$ is calculated as the factorized, weighted log probability over all generation steps:
\begin{equation}
\mathrm{Score}_{X \rightarrow Y} = \sum_{t=1}^{m}w_t\mathrm{log} P (y_t|X,\{y_{t^\prime}\}_{t^\prime=1}^{t-1}; \theta)
\label{eq:1}
\end{equation} 
where $w_t$ denotes the term importance score to put different emphasis on different target tokens $y_t$. BARTScore simply employs a uniform weighting scheme (all equal to 1). $\theta$ denotes the parameterized seq2seq model.

Our contributions consist of three modifications tailored to machine translation:

\medskip

\noindent\textbf{Data augmented translations.}
Unlike BARTScore, we employ M2M-100 \citep{fan2021beyond}, a non-English-centric multilingual machine translation system as our backbone seq2seq model, due to its superior performance.
As our main contribution, we translate the source (e.g., Src$_{fr}$ in Figure \ref{fig:model}) and the reference (Ref$_{en}$) into any languages as our augmented data (Trans1$_{xx}$ and Trans2$_{yy}$) for evaluating the hypothesis (Hypo$_{en}$).
In addition to the three directions covered by BARTScore, our metric takes into consideration all generation directions centered on the hypothesis connecting the source, the reference, and the two data augmented translations, i.e., in total 8 directions as the black (dashed and solid) arrows depicted in Figure \ref{fig:model}. DATScore is calculated as the weighted average of the scores associated with all the directions:
\begin{equation}
\mathrm{DATScore} = \sum_{X,Y} w_{X \rightarrow Y}\mathrm{Score}_{X \rightarrow Y} ; X \neq Y
\label{eq:2}
\end{equation}
where $w_{X \rightarrow Y}$ denotes the weight of the direction ${X \rightarrow Y}$, as detailed below.

\medskip

\noindent\textbf{One-vs-rest score averaging method.}
We observed empirically that sometimes, one direction score might strongly disagree with the others, likely being an outlier (failed evaluation).
This may significantly affect the final DATScore correlations with the human meta-evaluations, if a simple arithmetic averaging method is applied (like BARTScore in computing F-score).
To reduce this effect, we weigh each direction with the sum of the Pearson correlations of its scores with the scores of all the other directions:

\begin{multline}
w_{X \rightarrow Y} = \sum_{X',Y'} Corr(\mathrm{Score}_{X \rightarrow Y}, \mathrm{Score}_{X' \rightarrow Y'}) \\
\mathrm{s.t.} \quad (X, Y) \neq (X', Y')
\label{eq:3}
\end{multline}
This one-vs-rest method will assign a low weight to the direction score poorly correlated with the rest scores, thus reducing its negative effect on the averaging result.

\medskip

\noindent\textbf{Entropy-based term weighting scheme.}
BARTScore gives an equal weight $w_t$ to every token in Equation \ref{eq:1} (uniform weighting). 
Instead, we introduce a novel scheme to give different importance to different target tokens $y_t$, based on the entropy:
\begin{equation}
w_t = -\sum_{i=1}^{v} \mathrm{P_t}(z_{i}) \mathrm{logP_t}(z_{i})
\label{eq:4}
\end{equation}
where $v$ denotes the size of the output generation vocabulary. $\mathrm{P_t}(z_{i})$ represents the probability of the $i$-th token in the vocabulary at time step $t$.
We assume that when the model is very confident in generating the target token (low entropy), then this token is non-informative (e.g., stopword).
On the other hand, when the model is less confident (higher entropy), the target word is more informative, and then a higher weight should be assigned. 

\medskip

The effectiveness of all our choices regarding the above contributions is shown by our ablation studies (see Section \ref{sec:abla}).

\begin{table*}[!ht]
    \centering
    \small
    \def\arraystretch{1.4}
    \setlength{\tabcolsep}{2.9pt} 
    \begin{tabular}{l|rc|ccccccc|c}
    \hline
        Metric && Model & 
        \makecell{$|r|$:cs \shortrightarrow{.1cm}en \\ / \\ $\tau$:en \shortrightarrow{.1cm}cs \vspace{0.1cm}}&
        \makecell{$|r|$:de \shortrightarrow{.1cm}en \\ / \\ $\tau$:en \shortrightarrow{.1cm}de \vspace{0.1cm}}&
        \makecell{$|r|$:fi \shortrightarrow{.1cm}en \\ / \\ $\tau$:en \shortrightarrow{.1cm}fi \vspace{0.1cm}}&
        \makecell{$|r|$:lv \shortrightarrow{.1cm}en \\ / \\ $\tau$:en \shortrightarrow{.1cm}lv \vspace{0.1cm}}&
        \makecell{$|r|$:ru \shortrightarrow{.1cm}en \\ / \\ - \vspace{0.1cm}}&
        \makecell{$|r|$:tr \shortrightarrow{.1cm}en \\ / \\ $\tau$:en \shortrightarrow{.1cm}tr \vspace{0.1cm}}&
        \makecell{$|r|$:zh \shortrightarrow{.1cm}en \\ / \\ - \vspace{0.1cm}} & Avg. \\ \hline \hline
        BLEU &1a)& N/A & 34.4/22.0 & 36.6/23.6 & 44.4/42.1 & 32.1/21.5 & 41.3/- & 44.1/33.6 & 44.0/- & 37.8/27.3 \\ \hline
        BERTScore &1b)& RL/mBERT & 71.0/43.8 & \textbf{74.5}/40.4 & 83.3/58.8 & 75.6/46.6 & 74.6/- & 75.1/57.1 & 77.5/- & 75.9/49.3\\ \hline
        MoverScore &1c)& BB/mBERT & 66.6/38.3 &  70.6/35.9 & 82.2/54.2  & 71.7/37.8 & 73.7/- & 76.1/49.8 & 74.3/- & 73.6/43.2  \\ \hline
         \multirow{3}{4.5em}{BARTScore} &1d)& BL+para/mBART & 68.4/39.0 & 70.8/33.4 & 79.4/50.4 & 74.9/50.4 & 71.8/- & 73.9/53.8 & 76.0/- & 73.6/45.4 \\ \cline{2-11}
         &1e)& M2M-100\_418M & 65.9/45.0 & 66.1/44.5 & 79.9/59.2 & 71.7/40.3 & 69.0/- & 71.8/70.9 & 71.6/- & 
70.9/52.0 \\ 
         &1f)& M2M-100\_1.2B & 67.4/49.6 & 69.3/49.2 & 80.7/63.5 & 73.7/46.9 & 70.4/- & 71.6/\textbf{72.5} & 73.0/- & 72.3/56.3 \\ \hline \hline
         \multirow{2}{4.5em}{DATScore} &1g)& M2M-100\_418M & 68.6/51.1 & 68.5/48.1 & 82.0/63.7 & 74.7/48.3 & 73.0/- & 77.6/70.9 & 76.5/- & 74.4/56.4 \\ 
         &1h)& M2M-100\_1.2B & \textbf{71.3}/\textbf{53.9} & 72.9/\textbf{52.2} & \textbf{83.5}/\textbf{66.3} & \textbf{76.8}/\textbf{52.0} & \textbf{75.9}/- & \textbf{78.1}/70.9 & \textbf{77.7}/- &
 \textbf{76.6}/\textbf{59.1} \\ \hline

    \end{tabular}
    \caption{Absolute Pearson correlation ($|r|$) for \texttt{to-English} and Kendall correlations ($\tau$) for \texttt{from-English} with segment-level human scores on WMT17. BB stands of Bert-Base, RL for RoBERTa-Large and BL for BART-Large.}
    \label{tab:wmt17}
\end{table*}



\begin{table*}[!ht]
    \centering
    \small
    \def\arraystretch{1.4}
    \setlength{\tabcolsep}{3.2pt} 
    \begin{tabular}{l|rc|ccccccc|c}
    \hline
        Metric && Model & \makecell{$\tau$:cs \shortrightarrow{.1cm}en \\ / \\ $\tau$:en \shortrightarrow{.1cm}cs \vspace{0.1cm}}&
        \makecell{$\tau$:de \shortrightarrow{.1cm}en \\ / \\ $\tau$:en \shortrightarrow{.1cm}de \vspace{0.1cm}}&
        \makecell{$\tau$:et \shortrightarrow{.1cm}en \\ / \\ $\tau$:en \shortrightarrow{.1cm}et \vspace{0.1cm}}&
        \makecell{$\tau$:fi \shortrightarrow{.1cm}en \\ / \\ $\tau$:en \shortrightarrow{.1cm}fi \vspace{0.1cm}}&
        \makecell{$\tau$:ru \shortrightarrow{.1cm}en \\ / \\ $\tau$:en \shortrightarrow{.1cm}ru \vspace{0.1cm}}&
        \makecell{$\tau$:tr \shortrightarrow{.1cm}en \\ / \\ $\tau$:en \shortrightarrow{.1cm}tr \vspace{0.1cm}}&
        \makecell{$\tau$:zh \shortrightarrow{.1cm}en \\ / \\ $\tau$:en \shortrightarrow{.1cm}zh \vspace{0.1cm}}&
        Avg. \\ \hline \hline
        BLEU &2a)& N/A & 23.3/38.9 & 41.5/62.0 & 38.5/41.4 & 15.4/35.5 & 22.8/33.0 & 14.5/26.1 & 17.8/31.1 & 24.8/38.3 \\ \hline
        BERTScore &2b)& RL/mBERT &40.4/55.9 & \textbf{55.0}/72.7 & 39.7/58.4 & 29.6/53.9 & 35.3/42.4 & 29.2/38.9 & \textbf{26.4}/36.1 & 36.5/51.2 \\ \hline
        MoverScore &2c)& BB/mBERT & 36.8/44.6 & 53.9/68.4 & 39.4/52.7 & 28.7/50.9 & 27.9/40.1 & \textbf{33.6}/32.5 & 25.6/35.2 & 35.1/46.3\\ \hline
        \multirow{3}{4.5em}{BARTScore} &2d)& BL+para/mBART & 39.6/50.2 & 54.7/65.0 & 39.4/53.3 & 28.9/57.2 & 34.6/37.0 & 27.4/37.7 & 24.9/32.4 & 35.6/47.5 \\ \cline{2-11} 
         &2e)& M2M-100\_418M & 36.3/55.4 & 53.5/72.2 & 37.6/58.4 & 26.3/60.2 & 33.4/44.4 & 26.8/45.1 & 23.4/31.3 & 33.9/52.4  \\
         &2f)& M2M-100\_1.2B & 38.4/\textbf{63.5} & 54.6/\textbf{76.2} & 39.2/63.2 & 27.9/64.5 & 35.7/45.6 & 28.5/50.2 & 24.3/34.7 & 35.5/56.8 \\ \hline \hline
         \multirow{2}{4.5em}{DATScore} &2g)& M2M-100\_418M & 38.6/53.5 & 53.5/71.3 & 39.3/64.0 & 28.4/62.2 & 34.9/44.4 & 28.5/47.9 & 25.3/34.0 & 35.5/53.9 \\  
         &2h)& M2M-100\_1.2B & \textbf{40.7}/61.9 & 54.9/\textbf{76.2} & \textbf{40.5}/\textbf{68.2} & \textbf{30.4}/\textbf{67.9} & \textbf{36.4}/\textbf{46.2} & 31.0/\textbf{52.7} & 26.3/\textbf{36.6} & \textbf{37.2}/\textbf{58.5} \\ \hline

    \end{tabular}
    \caption{Kendall correlations ($\tau$) for \texttt{to-English} and \texttt{from-English} with segment-level human scores on WMT18. BB stands of Bert-Base, RL for RoBERTa-Large and BL for BART-Large.}
    \label{tab:wmt18}
\end{table*}

\section{Experiments}
\subsection{Experimental settings} \label{subsec:exp_settings}
We benchmark DATScore on two commonly used meta-evaluation datasets for machine translation metrics: WMT17 \cite{bojar-etal-2017-results} and WMT18 \cite{ma-etal-2018-results} consisting of multiple \texttt{to\_English} and \texttt{from\_English} language pairs.
For each pair, a few thousand examples are available, each being made of a \textit{source}, a \textit{reference}, a \textit{hypothesis} and a \textit{label} produced by human annotators, assessing the quality of the system generated \textit{hypothesis}.
Depending on the \textit{label} type, we use Kendall's Tau $\tau$ correlations or absolute Pearson $|r|$ correlations. The former is used when relative ranking is provided, and the latter in the case of direct assessment. We adopt the Kendall's Tau-like formulation proposed in \citep{bojar-etal-2017-results}:
\begin{equation}
\tau = \frac{|Concordant| - |Discordant|}{|Concordant| + |Discordant|}    
\end{equation}
where $|Concordant|$ is the number of examples on which the metric agrees with the human relative ranking, and $|Discordant|$ is the number of examples when they disagree. \\

To compute DATScore, two M2M-100 models: M2M-100\_{418M}\footnote{https://huggingface.co/facebook/m2m100\_418M} and M2M-100\_{1.2B}\footnote{https://huggingface.co/facebook/m2m100\_1.2B} are adopted (418M and 1.2B refer to the model sizes). They are finetuned to translate a source text to a target text by providing the source language code (e.g. "fr") at the beginning of the encoder input sequence, and a target language code at the beginning of the decoder input sequence.
In our experiments, when English is the target language (\texttt{to-English}), we choose English for Trans1 and Spanish for Trans2 (see Figure \ref{fig:model}). Otherwise, whenever English is the source language (\texttt{from-English}), we choose Spanish for Trans1 and English for Trans2. This choice is motivated by the fact that English and Spanish are the top two represented languages in the training set of M2M-100 \cite{fan2021beyond}. 

\subsection{Main results}

We compare the performance of our metric against BLEU and three other reference-based unsupervised metrics: BERTScore\footnote{https://github.com/Tiiiger/bert\_score}, MoverScore\footnote{https://github.com/AIPHES/emnlp19-moverscore} and BARTScore\footnote{https://github.com/neulab/BARTScore} (detailed in Subsection \ref{subsec:metrics} and Section \ref{sec:datscore}), using their official implementations.
Experimental results are reported in Table \ref{tab:wmt17} and \ref{tab:wmt18}.
Following their original settings, we use different underlying language models for each baseline metric. 
For BERTScore and MoverScore, RoBERTa-Large (RL; \citealp{liu2019roberta})  and Bert-Base (BB) are used respectively when we evaluate \texttt{to-English} translations, and mBERT \cite{devlin-etal-2019-bert} for \texttt{from-English} translations. 
In the case of BARTScore, we use a BART-Large (BL) checkpoint (finetuned on CNNDM \cite{see-etal-2017-get} and ParaBank2 \cite{hu-etal-2019-large} datasets) for evaluating \texttt{to-English} translations, and an mBART-50 model \cite{escolano-etal-2021-multilingual} for \texttt{from-English} translations.

Overall, results show that, on average, across all language pairs, DATScore significantly outperforms all 4 baseline metrics under their original model settings (rows 1a-1d and 2a-2d). 
Specifically, with respect to the best performing baseline BERTScore (row 1b and 2b), our metric provides a performance boost of 0.7 for \texttt{to-English} case and of 9.8 for \texttt{from-English} case on WMT17 dataset in Table \ref{tab:wmt17}, and achieves a gain of 0.7 and of 7.3 respectively on WMT18 dataset in Table \ref{tab:wmt18}.
These averaging results demonstrate the superiority and applicability of DATScore in evaluating general machine translations of many languages. Moreover, it is interesting to note that our improvement is much more significant in \texttt{from-English} case, which makes DATScore particularly well-suited to evaluate hypothesis translations in non-English languages, often with low resource. 
We hypothesize that this is due to the inconsistency of underlying language models. The baselines adopt a monolingual model for evaluating English, but a multilingual one for non-English languages. However, DATScore uses a single multilingual M2M-100 model for both cases. It is known that, in general, monolingual models outperform multilingual competitors. Thus, it is reasonable that when comparing multilingual-based DATScore against monolingual baselines in the \texttt{to-English} case, DATScore achieves a smaller improvement than in the other \texttt{from-English} case, where the comparison is fairer (multilingual vs. multilingual).

By looking across specific language pairs and directions, we observe DATScore constantly performs better than 4 baseline metrics with a few exceptions, i.e., de \shortrightarrow{.1cm}en (-1.6) in Table \ref{tab:wmt17}, and de \shortrightarrow{.1cm}en (-0.1), tr \shortrightarrow{.1cm}en (-2.6), and zh \shortrightarrow{.1cm}en (-0.1) in Table \ref{tab:wmt18}.
Despite these small drops in the performance, DATScore brings a larger margin of improvement in most cases, such as en \shortrightarrow{.1cm}tr up to 13.8 both on WMT17 and WMT18 datasets.

In the end, for the sake of having a complete comparison, we additionally evaluate BARTScore\footnote{The official implementation of BARTScore is slightly modified to take into account the languages tokens when using a multilingual model.} with M2M-100\_418M and M2M-100\_1.2B models (row 1e, 1f, 2e, and 2f) that are used as DATScore's underlying models. 
Results show that, only in the \texttt{from-English} case, while they bring an
improvement compared to the vanilla BARTScore
(row 1d and 2d), they are not able to yield as big of a gain as our metric, indicating that our achieved improvement is not solely due to the underlying language model, but also to taking additional generation directions into account, including those related to data augmented translations.


\begin{table}[t]
    \centering
    \small
     \def\arraystretch{1.3}
    \setlength{\tabcolsep}{2pt} 
    \begin{tabular}{l|c|ccc}
    \hline
        \multirow{2}{4.5em}{Metric} & \multirow{2}{2.5em}{Model} & \multicolumn{3}{c}{\underline{WebNLG}}  \\
        &  & SEMA & GRAM & FLU \\
        \hline
        \hline
        BLEU & N/A & 45.5 & 36.0 & 34.9 \\ \hline
        BERTScore & RoBERTa-Large & 56.1 & 60.8 & 54.8 \\ \hline
        MoverScore & BERT-Base & -9.9 & -27.8 & -20.6 \\ \hline
        \multirow{3}{4.5em}{BARTScore} & BART-Large+para & \textbf{71.9} & 61.3 & 57.4 \\ \cline{2-5} 
         & M2M-100\_418M & 64.9 & 62.8 & 56.0 \\
         & M2M-100\_1.2B & 66.1 & 63.9 & 57.2 \\ \hline
         \multirow{2}{4.5em}{DATScore} 
          & M2M-100\_418M & 69.9 & 62.9 & 57.2 \\ 
         & M2M-100\_1.2B & 70.4 & \textbf{63.7} & \textbf{57.9} \\
        \hline

    \end{tabular}
    \caption{Pearson correlation results on WebNLG dataset.}
    \label{tab:other_webnlg}
    
\end{table}

\begin{table}[t]
    \centering
    \scriptsize
     \def\arraystretch{1.3}
    \setlength{\tabcolsep}{2pt} 
    \begin{tabular}{l|c|c|cccc}
    \hline
        \multirow{2}{4.5em}{Metric} & \multirow{2}{2.5em}{Model} & \underline{REALSumm} & \multicolumn{4}{c}{\underline{SummEval}}  \\
         & & COV & COH & CONS & FLU & REL \\ 
        \hline
        \hline
        BLEU & N/A &  37.9 & 11.8 & 6.3 & 7.7 & 18.6 \\ \hline
        BERTScore & RoRERTa-Large & 41.2 & \textbf{33.9} & 10.5 & \textbf{15.0} & \textbf{35.9}\\ \hline
        MoverScore & BERT-Base & 44.1 & 14.4 & \textbf{14.7} & 13.8 & 29.1 \\ \hline
        \multirow{3}{4.5em}{BARTScore} & BART-Large+para & 31.7 & 20.8 & -3.5 & 6.7 & 22.2  \\  \cline{2-7} 
         & M2M-100\_418M & 30.1 & 14.8  & -2.3 & 3.0 & 19.8 \\
         & M2M-100\_1.2B & 32.0 & 17.1 & 1.1 & 6.7  & 22.8  \\ \hline
         \multirow{2}{4.5em}{DATScore} 
          & M2M-100\_418M & 44.7 & 17.1 & 4.4 & 4.6 & 26.3 \\ 
         & M2M-100\_1.2B & \textbf{45.5} & 19.5 & 6.8 & 8.2 & 30.2 \\
        \hline

    \end{tabular}
    \caption{Pearson correlation results on two summarization datasets: REALSumm and SummEval.}
    \label{tab:other_summarization}
    
\end{table}

\section{Other NLG tasks}
In addition to machine translation, our main focus, we evaluate DATScore on other NLG tasks, including data-to-text generation, abstractive summarization, and image captioning. 
To work around the different modalities of source inputs represented in these tasks (e.g., not able to create a data augmented translation with an image), we adapt DATScore to only consider 4 generation directions: \textit{Hypo$\leftrightarrow$Ref} and \textit{Hypo$\leftrightarrow$Trans2}.

\medskip

\noindent\textbf{Data-to-text.} Table \ref{tab:other_webnlg} shows the performance of DATScore compared to the other baselines on the WebNLG data-to-text dataset \cite{shimorina2018webnlg}, which contains 2000 descriptions of structured tables along with their corresponding references. In addition, human assessments covering three dimensions are provided (\textit{semantics}, \textit{grammar}, and \textit{fluency}). The results show that DATScore significantly outperforms all the other metrics in two settings (grammar and fluency) out of three, while being very competitive in the third setting (semantics). Surprisingly, BERTScore is largely behind DATScore, and MoverScore failed to correlate positively with human judgments in all dimensions.

\medskip

\noindent\textbf{Summarization.} Table \ref{tab:other_summarization} shows the evaluation of the different metrics on two summarization meta-evaluation datasets: REALSumm \cite{bhandari-etal-2020-evaluating} and SummEval \cite{fabbri-etal-2021-summeval}. Both datasets contain a few thousand examples of system-generated summaries and their references. The generated summaries are annotated with \textit{lightweight pyramids} \cite{shapira-etal-2019-crowdsourcing} method in the case of REALSumm, while the annotations in SummEval cover four dimensions: \textit{coherence}, \textit{consistency}, \textit{fluency}, and \textit{relevance}. On REALSumm, DATScore has the best performance compared to all the other baselines even when using its smaller version (M2M-100\_418M). However, despite its higher correlations compared to BARTScore and MoverScore, DATScore fails to outperform BERTScore on the different dimensions of SummEval.

\medskip

\noindent\textbf{Image captioning.} We consider Flickr8K \cite{hodosh2013framing} and PASCAL-50S \cite{vedantam2015cider}, two image captioning datasets. The former is annotated with scores from 1 to 4 assessing the relevance of the captions, and the latter is annotated with relative ranking (i.e., given two descriptions which one is better). Table \ref{tab:other_image_captioning} shows that in this task, DATScore is competitive to BARTScore and BERTScore. Surprisingly, MoverScore significantly outperforms all the other metrics despite its poor performance on the other datasets.

\medskip

Finally, although not the top-performing metric across all tasks, DATScore showed an overall stable and competitive performance. Conversely, each of the other metrics fails in evaluating generations, at least in one of the tasks.
For example, BERTScore and MoverScore have poor performance on the WebNLG dataset. On the other hand, although BARTScore is finetuned on an abstractive summarization dataset, it fails to correlate well with human judgment on REALSumm and SummEval. This finding suggests that DATScore can be safely used to evaluate NLG systems in other tasks for different evaluation dimensions, regardless of being initially designed for machine translation evaluation.

\begin{table}[t]
    \centering
    \small
     \def\arraystretch{1.3}
    \setlength{\tabcolsep}{2pt} 
    \begin{tabular}{l|c|c|c}
    \hline
        \multirow{2}{4.5em}{Metric} & \multirow{2}{2.5em}{Model} & \underline{Flickr8K} & \underline{PASCAL-50S}  \\
         & & RELE & RR \\
        \hline
        \hline
        BLEU & N/A & 13.8 & 8.1 \\ \hline
        BERTScore & RoBERTa-Large & 46.1 & \textbf{33.8} \\ \hline
        MoverScore & BERT-Base & \textbf{52.5} & 33.2 \\ \hline
        \multirow{3}{4.5em}{BARTScore} &BART-Large+para & 44.8 & 33.1  \\
        \cline{2-4} 
         & M2M-100\_418M & 34.3 & 29.6 \\
         & M2M-100\_1.2B & 34.6 & 26.3 \\ \hline
         \multirow{2}{4.5em}{DATScore} 
          & M2M-100\_418M & 42.6 & 29.6 \\ 
         & M2M-100\_1.2B & 45.3 & 31.4 \\
        \hline
        
    \end{tabular}
    \caption{Pearson correlation Results on two Image Captioning datasets: Flickr8K and PASCAL-50S.}
    \label{tab:other_image_captioning}
\end{table}

\begin{table}[t]
    \centering
    \small
     \def\arraystretch{1.5}
    \setlength{\tabcolsep}{2pt} 
    \begin{tabular}{|c|c|c|c|}
    \hline
        \makecell{Entropy-based \\ weighting} & \makecell{One-vs-rest \\ weighting} & \texttt{to\_English} & \texttt{from\_English} \\
        \hline
        \hline
        \cmark & \cmark & \textbf{37.2} & \textbf{58.5} \\
        \cmark & \xmark & 37.1 & 58.1 \\
        \xmark & \cmark & 36.4 & 55.9 \\
        \xmark & \xmark & 36.4 & 56.0 \\
        \hline

    \end{tabular}
    \caption{The average Kendall correlation (to/from)-English when the entropy-based and one-vs-rest weighting are included or excluded. Experiments are conducted on WMT18.}
    \label{tab:ablation_weighting}
\end{table}

\begin{figure*}[ht]
  \includegraphics[scale=0.54]{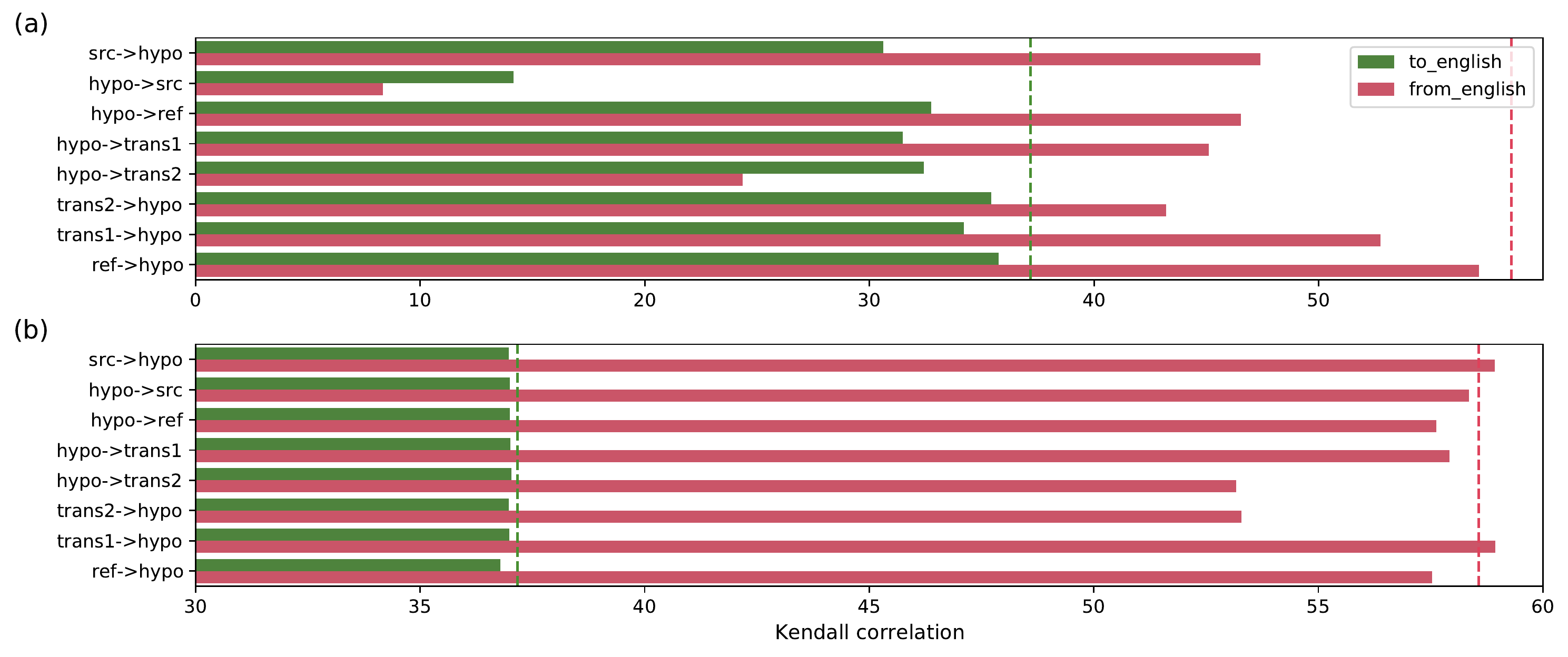}
  \caption{(a): The horizontal bars represent the Kendall correlations of \textbf{each individual generation direction}. (b): The horizontal bar represents the Kendall correlation of \textbf{a variant of DATScore with excluding the single generation direction} of the line. Both in (a) and (b), the dashed vertical lines represent the Kendall correlation of the vanilla and \textbf{complete DATScore}. Correlation results of \texttt{to-English} (in green) and \texttt{from-English} (in red) cases are calculated w.r.t human judgments, and averaged over all languages pairs. Experiments are conducted on WMT18.}
  \label{fig:ablation_directions}
\end{figure*}

\section{Ablation study} \label{sec:abla}

To validate our different choices with regard to DATScore, we conducted ablation studies on:\\
1) the contributions of all 8 direction scores, results are illustrated in Figure \ref{fig:ablation_directions}. \\2) the effectiveness of our \textit{one-vs-rest} score averaging and \textit{entropy-based} term weighting strategies (See Section \ref{sec:datscore}), results are reported in Table \ref{tab:ablation_weighting}.

\medskip

\noindent\textbf{Contributions of all direction scores.}
From Figure \ref{fig:ablation_directions}(a), we observe that none of the individual directions (horizontal bars) has a better correlation with human judgments than DATScore (dashed vertical lines), which confirms the importance of our ensemble approach. In Figure \ref{fig:ablation_directions}(b), we can see that all variants excluding one direction will lead, in almost all cases, to a drop in the performance, compared to the complete DATScore in which all directions are included. 
Besides, in the case of \texttt{to-English} translations, we can see that the drop in the performance is almost the same for all exclusions of direction. While for \texttt{from-English} translations, the largest drop in performance is observed when \textit{Hypo$\rightarrow$Trans2} and \textit{Trans2$\rightarrow$Hypo} are excluded. This finding highlights the important contribution of our augmented data, especially in the low resource language settings (\texttt{from-English}). In the end, we can see that excluding \textit{Src$\rightarrow$Hypo} or \textit{Trans1$\rightarrow$Hypo} directions can lead to a slightly better final score. We leave the investigation of the potential negative impact of the two directions to future work.

\medskip

\noindent\textbf{One-vs-rest and entropy-based weighting strategies.}
Table \ref{tab:ablation_weighting} shows the performance of DATScore variants with respect to different combinations of applying or not our proposed weighting strategies. Note that when \textit{one-vs-rest} and \textit{entropy-based} weightings are not applied, they are replaced with a simple uniform averaging approach (as used in BARTScore). A performance drop is observed when excluding one of the two weighting strategies, especially for the entropy-based method, whose inclusion leads to an improvement of 2.5 compared to the uniform weighting. This experiment confirms the positive impact of our proposed weighting methods and motivates future work further to investigate a more elaborated approach in this direction.


\section{Conclusion}
In this work, we proposed one of the first applications of data augmentation techniques to NLG evaluation. To obtain an evaluation score of the translation hypothesis, our developed metric DATScore additionally leverages newly translated copies augmented from the source and reference texts. We also proposed two novel strategies for score averaging and term weighting to improve the original, naive score computing process of BARTScore, on the basis of which our work is built.
Experimental results show that DATScore achieved a higher correlation with human meta-evaluations, in comparison with the other recent state-of-the-art metrics, especially for those less represented languages other than English. Moreover, ablation studies show the effectiveness of our newly proposed score computing approaches, and extended experiments showed an overall stable and competitive performance of DATScore on more NLG tasks.

\section*{Limitations}

In this section, we list some limitations that are worth further investigation in future works:

\medskip

\noindent 1) DATScore requires generating additional data augmented translations to perform the evaluation. This process might be time-consuming depending on the adopted backbone seq2seq model, especially if the original text is long. Thus, the performance scalability can be investigated in future complementary experiments.

\medskip

\noindent 2) We chose to use English and Spanish to create data augmented translations for the reason that they are the two most represented languages in the training of the M2M-100 model (see Subsection \ref{subsec:exp_settings}). However, this leaves a question about the performance of DATScore with augmentations varying in other languages (e.g., Chinese).
Moreover, for the sake of simplicity, we decided only to include a single translated copy of the source text and the reference text. However, this can be easily extended, and more augmented translations can be created in more languages.
We expect to see an improvement in performance with diminishing returns.

\medskip

\noindent 3) BARTScore only considers the 8 generation directions centered on the hypothesis connecting with the source, the reference, and the two data augmented translations (see Section \ref{sec:datscore}). However, other connections exist between these entities, such as \textit{Src$\rightarrow$Ref} and \textit{Trans1$\rightarrow$Src} (see Figure \ref{fig:model}).
Therefore, future research could be dedicated to discovering the effect of these other directions and potentially leveraging them to improve the performance of DATScore.

\medskip

\noindent 4) Since our focus was on evaluating machine translation, we naturally chose translation for augmenting the data.
However, other data augmentation techniques could seamlessly integrate into DATScore, such as using a text paraphrasing model \citep{bandel-etal-2022-quality}.

\bibliography{aclanthology,googlescholar}
\bibliographystyle{acl_natbib}



\end{document}